%% file: iclr2027_conference.tex
\documentclass{article} %
\usepackage{iclr2027_conference,times}

\input{math_commands.tex}

\usepackage{hyperref}
\hypersetup{citebordercolor={0 0.5 0},linkbordercolor={0.8 0 0},urlbordercolor={0 0.5 0.8}}
\usepackage{url}
\expandafter\def\expandafter\UrlBreaks\expandafter{\UrlBreaks\do\/\do\-\do\_\do\.%
  \do\a\do\b\do\c\do\d\do\e\do\f\do\g\do\h\do\i\do\j\do\k\do\l\do\m\do\n\do\o%
  \do\p\do\q\do\r\do\s\do\t\do\u\do\v\do\w\do\x\do\y\do\z}
\usepackage{booktabs}
\usepackage{amssymb}
\usepackage{mathtools}
\usepackage[font=small,labelfont=bf]{caption}
\usepackage{graphicx}
\usepackage{xcolor}
\usepackage{colortbl}
\usepackage{tikz}
\usepackage{float}
\usepackage{placeins}
\usetikzlibrary{arrows.meta,positioning}

\title{Retrofitting Recurrent Vision--Language\\ Models from Pretrained Backbones}

\author{Madhav Iyengar$^{1}$ \quad Andreas Geiger$^{1,2}$ \quad Hilde K\"uhne$^{1}$\\[3pt]
$^{1}$University of T\"ubingen, T\"ubingen AI Center \quad $^{2}$KE:SAI\\[3pt]
Project page: \url{https://madhavesdios.github.io/recurrent-vlms}}

\iclrfinalcopy
\begin{document}

\maketitle
\ificlrfinal\lhead{Preprint. Under review.}\fi%
\ificlrfinal\hypersetup{pdftitle={Retrofitting Recurrent Vision-Language Models from Pretrained Backbones},pdfauthor={Madhav Iyengar, Andreas Geiger, Hilde Kühne}}\fi%

\begin{abstract}
Recent work shows that depth recurrence can be retrofitted into pretrained language models.
In this work, we adapt this paradigm for visual understanding, using depth recurrence to reduce the parameter footprint of a pretrained vision-language model (VLM). 
To this end, we prune layers from the language backbone of the VLM and loop over a subset of the remaining layers.
At each step of the loop, a linear adapter merges the block's output with the token representations from the preceding layers and projects the fused representation back into the block.
As the pruning impacts the original model performance, we propose a three-stage recovery training: first, we leverage a language-only pretraining to recover the performance of the language backbone, followed by an alignment of the vision-language projector, and a full fine-tuning of the model via visual instruction tuning. In the first and the last stage, we distill the outputs of the pretrained non-recurrent model into the recurrent model, which further improves the recovery.  
We evaluate the proposed approach on the TinyLLaVA-1.4B backbone without introducing any new vision-language data during retrofitting. 
Our experiments demonstrate that the retrofitted recurrent model outperforms the unpruned baseline trained with the same recovery data while storing 23.3\% fewer language parameters.  
We further show that the proposed retrofitting recurrence outperforms the setup of converting a recurrent LM into a VLM on the same data. 
Additionally, our ablations show how design choices, such as the adapter between recurrences and a larger recurrent block, lead to the improved performance of our approach.
Finally, we find that our proposed model gains on reasoning-oriented benchmarks, whereas its performance on basic visual understanding benchmarks remains on par with the baseline.

\end{abstract}

\input{sections/01_intro}

\input{sections/02_related}

\input{sections/03_method}
\input{sections/04_setup}

\input{sections/05_results}

\input{sections/07_discussion}

\newpage%
\ificlrfinal
\subsubsection*{Acknowledgments}
This work was funded by the Federal Ministry of Research, Technology and Space (BMFTR) under the project XEI, FKZ: 16IS24079A. The authors gratefully acknowledge the Gauss Centre for Supercomputing e.V.\ (www.gauss-centre.eu) for providing computing time on the JUPITER Booster at J\"ulich Supercomputing Centre (JSC) under the project token-compression, and the computing resources provided by the ML Cloud of the T\"ubingen AI Center. We are grateful to Jonas Geiping and Takeru Miyato for their insightful discussions on looped models.
\fi
\subsubsection*{Reproducibility statement}
Appendix~\ref{app:protocol} lists every training hyperparameter of every model in the
paper. We will release the code and the checkpoints.

\subsubsection*{AI use statement}
Generative-AI assistants were used to polish the writing and to implement the baselines
and the proposed method. 
The experiments, the results and their interpretation are the
authors' own.

\bibliography{references}
\bibliographystyle{iclr2027_conference}

\appendix
\input{sections/appendix}

\end{document}

%% file: math_commands.tex
\usepackage{amsmath,amsfonts,bm}

\def\eqref#1{equation~\ref{#1}}

\def\1{\bm{1}}

\DeclareMathAlphabet{\mathsfit}{\encodingdefault}{\sfdefault}{m}{sl}
\SetMathAlphabet{\mathsfit}{bold}{\encodingdefault}{\sfdefault}{bx}{n}

%% file: sections/01_intro.tex
\vspace{-6pt}%
\section{Introduction}\label{sec:intro}
\vspace{-5pt}%

\looseness = -1
Existing approaches for test-time compute scaling in vision-language models comprise chain-of-thought traces, latent reasoning tokens, searching over candidate answers, and test-time reinforcement learning \citep{mmcot2023,covt2025,llavacot2024,ttrv2025,ttssurvey2026}. These methods usually come with higher inference costs as each generated token requires a full forward pass and expands the context window.
To address this limitation, \emph{depth recurrence} \citep{geiping2025huginn,mcleish2025retrofitted} has been explored as an alternative for test-time scaling in language models. Depth recurrence allocates test-time compute to iterating a block of layers rather than generating tokens, thereby maintaining smaller context sizes and a smaller memory overhead independent of reasoning length. 

But depth recurrence has also gained traction in visual understanding. Here, recurrent VLM architectures are built in two ways: first, either by starting from a recurrent language model and post-hoc adding a vision backbone, or, second, by looping an existing VLM without removing any of its layers. The first approach is i.a. adopted by HIVE \citep{hive2026}, which extends a recurrent language model pretrained from scratch by a vision backbone. Examples for the second approach are, RecursiveVLM \citep{recursivevlm2026} and ReLoop-UME
\citep{reloopume2026}, which loop layers in pretrained VLMs without pruning, with a near-constant number of parameters.

\begin{figure}[t]
\centering
\input{figures/fig_overview.tikz}
\caption{\textbf{Prune-then-loop retrofit.} Left: a pretrained VLM embeds the image as vision tokens, the prompt as text tokens and passes them through its language backbone. The layers marked in grey are pruned. 
Right: Our retrofitting mechanism reuses the remaining layers. The prelude is executed once, and an adapter merges the prelude's output with the hidden state of the recurrent block initialized with noise. The adapter and recurrent block are applied repeatedly, with the coda used to decode to logits.}
\label{fig:overview}
\vspace{-8pt}
\end{figure}
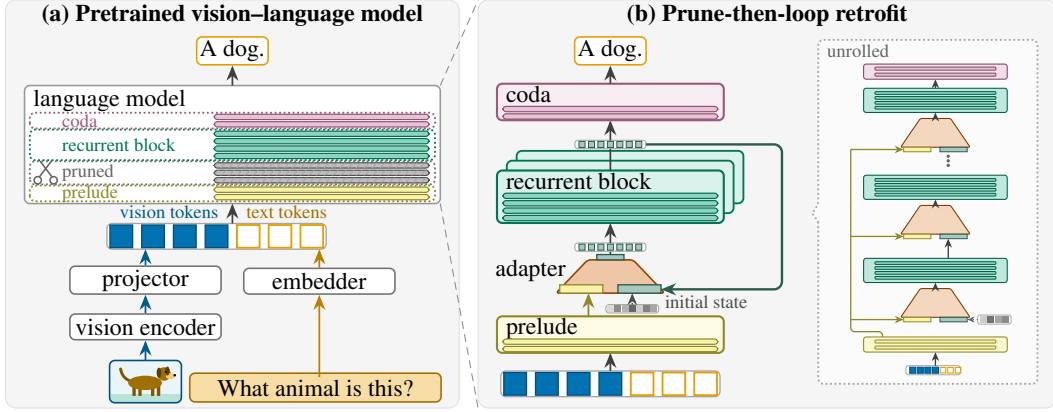

In contrast to these two regimes, we pursue a third option of retrofitting a pretrained VLM with recurrence, following a prune-then-loop strategy. Namely, we remove some intermediate blocks of the language backbone of the VLM and loop over a subset of the remaining layers. This strategy is supported by two complementary research findings: first, one set of recent works shows that a fraction of a VLM's layers can be removed at a limited cost in accuracy~\citep{huang2026vlmprune,interlace2025,shortlvlm2025},
and, second, another set of works shows that the layers of a pretrained model can be skipped or repeated at inference~\citep{polar2026,trainingfree2026}. We thus hypothesize that iteratively executing the subset of layers of a pruned
VLM can recover the capacity lost due to pruning. We test this hypothesis by
\emph{retrofitting}, i.e., by converting a pretrained non-recurrent VLM into
a recurrent one while reusing its existing weights rather than pretraining a
recurrent model from scratch.
We refer to the resulting model as retrofitted recurrent VLM (RR-VLM).
We prune a block of intermediate layers from the language backbone of a pretrained VLM and designate a contiguous subset of the remaining layers as a recurrent block
\citep{mcleish2025retrofitted}. To
feed the block's output back into its input for successive iterations, we
concatenate the output with the tokens
from the layers preceding the recurrent block and train a linear layer to
compress this concatenated representation to its original dimensionality
\citep{mcleish2025retrofitted}. 
An overview of the process is given in Figure~\ref{fig:overview}.
This surgery, however, degrades the pretrained language backbone, and although
text-only training repairs it, the repair leaves the language model misaligned
with the original projector. We resolve this with a three-stage training curriculum: first, we apply a text-only training to repair the backbone, second, we apply a projector-only training to restore the
image-text alignment, and third, we fine-tune the entire model on visual
instruction data. In the text-only and the instruction-tuning stage, we add a knowledge distillation (KD) term, with the pretrained model as teacher, to avoid the language backbone from drifting too far from the original representation that the projector was aligned with.

We evaluate the proposed method on the TinyLLaVA-1.4B backbone
\citep{tinyllava2024}, training it on 60B tokens of text as well as 493k image--caption
pairs and 660k instruction tuning samples from the base model's training data, without introducing any new vision-language data. We compare the resulting model against the pretrained TinyLLaVA-1.4B model and four baselines trained on the same data: (1) the unpruned model subjected to the identical recovery training, (2) the unpruned model without the text-only stage, (3) the pruned model without recurrence, and (4) a recurrent language model that has been extended to a VLM. 
Our experiments show that our proposed retrofitted recurrent model outperforms the unpruned baseline while having $23.3\%$ fewer language-model parameters and show that this gain is due to the recurrence rather than the recovery training. The gain concentrates on reasoning-oriented benchmarks, whereas performance on basic visual understanding benchmarks remains on par with the baseline. Compared to a recurrent language model extended to a VLM, our retrofitted model achieves a higher average accuracy. 
We further ablate each component of the retrofit: the training stages, the layer allocation, the knowledge distillation term, KV-cache sharing at inference and the adapter. We find that the text-only stage $S_0$ is necessary for the recovery, as removing it costs $3.68$ pp, primarily on the reasoning-oriented benchmarks.
Specifically, we evaluate our linear adapter against three variants: a gated adapter, the Parcae adapter \citep{parcae2026}, which multiplies the recurrent state with a learned per-channel factor within $(0,1)$ so that the hidden state remains bounded across recurrences, and no adapter at all. 
Our ablations show that, among the adapter variants, the linear adapter yields the best performance, whereas the gated adapter and the Parcae adapter score below the variant without an adapter. 
Furthermore, we show that at equal compute, a larger recurrent block results in better performance. 
We also show that distillation from the pretrained model improves recovery training and increases the performance of the retrofitted recurrent model. Finally, we find that a model trained at a fixed depth only works at the depth it was trained at, %
whereas a model trained with sampled recurrence works at other depths, but its performance does not improve beyond the trained depth. 
We also find that sharing the KV cache across recurrences during inference costs accuracy, which contrasts with the results reported for a model pretrained with recurrence \citep{geiping2025huginn}.

We summarize our contributions as follows: (1) We introduce the first prune-then-loop retrofit of a pretrained VLM evaluated on VQA, supported by a three-stage curriculum that repairs the pruned language backbone and restores projector alignment using the base model's vision-language training data. 
(2) We demonstrate that retrofitting recurrence directly into a pretrained VLM outperforms the alternative of adding a vision encoder to a pretrained recurrent language model. 
(3) We perform an extensive evaluation of the retrofitted recurrent VLM and quantify the impact of each design decision. %

%% file: figures/fig_overview.tikz
\definecolor{oiblue}{HTML}{0072B2}%
\definecolor{oiorange}{HTML}{E69F00}%
\definecolor{oiverm}{HTML}{D55E00}%
\definecolor{oigreen}{HTML}{009E73}%
\definecolor{oipurple}{HTML}{CC79A7}%
\definecolor{oiyellow}{HTML}{F0E442}%
\definecolor{oigrey}{HTML}{6E6E6E}%
\colorlet{oistate}{oigreen!45!oigrey}%
\usetikzlibrary{patterns}
\begin{tikzpicture}[
  x=1cm, y=1cm,
  >={Stealth[length=2.0mm,width=1.6mm]},
  font=\footnotesize,
  line width=0.55pt,
  panel/.style={fill=black!4, draw=black!12, line width=0.4pt, rounded corners=4pt},
  ptitle/.style={font=\small\bfseries, anchor=north, inner sep=0pt},
  lbl/.style={font=\footnotesize, anchor=north, inner sep=1pt},
  tag/.style={font=\scriptsize, inner sep=1pt},
  el/.style={rounded corners=2pt, line width=0.6pt, align=center, inner sep=1.6pt,
             minimum height=3.8mm},
  vis/.style={el, fill=oiblue!35, draw=oiblue!70!black},
  txt/.style={el, fill=oiorange!35, draw=oiorange!70!black},
  neutral/.style={el, fill=white, draw=black!55},
  vtk/.style={draw=oiblue!70!black, fill=oiblue, line width=0.5pt,
              minimum size=3.0mm, inner sep=0pt},
  ttk/.style={draw=oiorange, fill=white, line width=0.7pt,
              minimum size=3.0mm, inner sep=0pt},
  strip/.style={draw=black!35, fill=white, line width=0.5pt, rounded corners=2pt},
  bar/.style={rounded corners=1.2pt, line width=0.45pt},
  bpre/.style={bar, fill=oiyellow!45, draw=oiyellow!60!black},
  brem/.style={bar, fill=oigrey!45, draw=oigrey!70!black},
  bblk/.style={bar, fill=oigreen!45, draw=oigreen!70!black},
  bcod/.style={bar, fill=oipurple!45, draw=oipurple!70!black},
  gdot/.style={densely dotted, line width=0.7pt, rounded corners=2pt},
  grp/.style={rounded corners=2pt, line width=0.8pt},
  gpre/.style={grp, fill=oiyellow!14, draw=oiyellow!60!black},
  gblk/.style={grp, fill=oigreen!14, draw=oigreen!75!black},
  gcod/.style={grp, fill=oipurple!14, draw=oipurple!75!black},
  ybar/.style={line width=0.5pt, fill=oiyellow!45, draw=oiyellow!60!black},
  sbar/.style={line width=0.5pt, fill=oistate!45, draw=oistate!70!black},
  fl/.style={->, line width=0.65pt, draw=black!75},
  fv/.style={fl, draw=oiblue!80!black},
  ft/.style={fl, draw=oiorange!80!black},
  mod/.style={el, fill=white, draw=black!55, minimum height=3.4mm, inner sep=1.2pt,
             minimum width=2.0cm},
  ans/.style={el, fill=white, draw=oiorange, minimum height=3.4mm, inner sep=1.2pt},
  mfl/.style={->, >={Stealth[length=1.2mm,width=1.0mm]}, line width=0.5pt, draw=black!75},
  mybus/.style={line width=0.5pt, draw=oiyellow!60!black, rounded corners=1.5pt},
  mdot/.style={->, >={Stealth[length=1.0mm,width=0.8mm]}, line width=0.4pt, draw=black!60, densely dotted},
  mpre/.style={fill=oiyellow!25, draw=oiyellow!60!black, line width=0.4pt, rounded corners=1pt},
  mad/.style={fill=oiverm!25, draw=oiverm!60!black, line width=0.4pt, rounded corners=0.6pt},
  mblk/.style={fill=oigreen!25, draw=oigreen!60!black, line width=0.4pt, rounded corners=1pt},
  mcod/.style={fill=oipurple!25, draw=oipurple!60!black, line width=0.4pt, rounded corners=1pt},
  ftitle/.style={font=\footnotesize, anchor=north, inner sep=0pt,
                 text height=2.0mm, text depth=0.6mm},
]

\def\fovtokstrip#1#2{%
  \draw[strip] ({#1-0.19},{#2-0.17}) rectangle ({#1+2.71},{#2+0.17});
  \foreach \q in {0,...,3} { \node[vtk] at ({#1+0.42*\q},#2) {}; }
  \foreach \q in {4,5,6} { \node[ttk] at ({#1+0.42*\q},#2) {}; }
}
\def\fovstatestrip#1#2{%
  \draw[strip, line width=0.4pt, rounded corners=1.2pt] ({#1-0.47},{#2-0.055}) rectangle ({#1+0.47},{#2+0.055});
  \foreach \q in {-3,...,3}
    \draw[line width=0.35pt, fill=oistate!45, draw=oistate!70!black]
         ({#1+0.125*\q-0.04},{#2-0.04}) rectangle ({#1+0.125*\q+0.04},{#2+0.04});
}

\draw[panel] (0.00,0.03) rectangle (6.00,5.43);
\draw[panel] (6.25,0.03) rectangle (13.90,5.43);
\node[ptitle] at (3.000,5.38) {(a) Pretrained vision--language model};
\node[ptitle] at (10.075,5.38) {(b) Prune-then-loop retrofit};

\def\fovcL{1.85}\def\fovcR{4.15}%
\begin{scope}[shift={(\fovcL-0.475,0.10)}, scale=1.1176]
  \def\dx{0}\def\dy{0}%
  \begin{scope}
    \clip[rounded corners=2pt] (\dx,\dy) rectangle ++(0.85,0.50);
    \fill[oiblue!8] (\dx,\dy) rectangle ++(0.85,0.50);
    \fill[oigreen!28] (\dx,\dy) rectangle ++(0.85,0.10);
    \draw[oiorange!55!black, line width=1.1pt, line cap=round]
         (\dx+0.24,\dy+0.26) .. controls (\dx+0.16,\dy+0.29) and (\dx+0.14,\dy+0.35) .. (\dx+0.13,\dy+0.40);
    \foreach \lx in {0.27,0.34,0.49,0.56}
      \fill[oiorange!55!black, rounded corners=0.3pt] (\dx+\lx,\dy+0.08) rectangle ++(0.04,0.13);
    \fill[oiorange!70!black] (\dx+0.42,\dy+0.235) ellipse [x radius=0.20, y radius=0.085];
    \fill[oiorange!70!black] (\dx+0.645,\dy+0.34) circle (0.08);
    \fill[oiorange!55] (\dx+0.72,\dy+0.315) ellipse [x radius=0.055, y radius=0.036];
    \fill[oiorange!40!black] (\dx+0.585,\dy+0.355) ellipse [x radius=0.028, y radius=0.06];
    \fill[oiorange!40!black] (\dx+0.685,\dy+0.375) ellipse [x radius=0.026, y radius=0.053];
    \fill[black] (\dx+0.665,\dy+0.365) circle (0.012);
    \fill[black] (\dx+0.772,\dy+0.325) circle (0.015);
  \end{scope}
  \draw[oiblue!70!black, line width=0.6pt, rounded corners=2pt] (\dx,\dy) rectangle ++(0.85,0.50);
\end{scope}
\node[txt, anchor=south west, minimum width=3.325cm] (prompt) at (2.425,0.10) {What animal is this?};
\node[mod, anchor=south] (enc)  at (\fovcL,0.923) {vision encoder};
\node[mod, anchor=south] (proj) at (\fovcL,1.527) {projector};
\node[mod, anchor=south] (emb)  at (\fovcR,1.527) {embedder};
\draw[fv] (\fovcL,0.659) -- (enc.south);
\draw[fv] (enc.north) -- (proj.south);
\draw[ft] (\fovcR,0.44) -- (emb.south);
\fovtokstrip{1.53}{2.301}
\draw[fv] (proj.north) -- (\fovcL,2.131);
\draw[ft] (emb.north) -- (\fovcR,2.131);
\node[tag, text=oiblue!80!black, anchor=south east] at (2.86,2.491) {vision tokens};
\node[tag, text=oiorange!70!black, anchor=south west] at (3.14,2.491) {text tokens};
\def\fovLb{2.735}\def\fovLt{4.296}%
\draw[fill=white, draw=black!40, line width=0.6pt, rounded corners=2pt]
     (0.25,\fovLb) rectangle (5.75,\fovLt);
\draw[fl] (3.00,2.471) -- (3.00,\fovLb);
\node[anchor=north west, inner sep=0pt] at (0.36,\fovLt-0.05) {language model};
\def\fovbx{2.76}%
\draw[gdot, draw=oiyellow!60!black] (0.32,2.765) rectangle (5.68,2.973);
\foreach \y in {2.820,2.918} \draw[bpre] (\fovbx,\y-0.035) rectangle (5.60,\y+0.035);
\node[tag, inner sep=0pt, text=oiyellow!60!black, anchor=west] at (0.74,2.869) {prelude};
\draw[gdot, draw=oigrey] (0.32,2.988) rectangle (5.68,3.294);
\foreach \y in {3.043,3.141,3.239}
  \draw[brem, postaction={pattern=north east lines, pattern color=black!35}] (\fovbx,\y-0.035) rectangle (5.60,\y+0.035);
\begin{scope}[shift={(0.53,3.141)}, draw=black!60, line width=0.6pt]
  \draw (-0.085,-0.055) -- (0.075,0.155);
  \draw (0.085,-0.055) -- (-0.075,0.155);
  \draw (-0.105,-0.098) circle (0.048);
  \draw (0.105,-0.098) circle (0.048);
\end{scope}
\node[tag, inner sep=0pt, text=black!60, anchor=west] at (0.74,3.141) {pruned};
\draw[gdot, draw=oigreen!75!black] (0.32,3.309) rectangle (5.68,3.713);
\foreach \y in {3.364,3.462,3.560,3.658} \draw[bblk] (\fovbx,\y-0.035) rectangle (5.60,\y+0.035);
\node[tag, inner sep=0pt, text=oigreen!75!black, anchor=west] at (0.74,3.511) {recurrent block};
\draw[gdot, draw=oipurple!75!black] (0.32,3.728) rectangle (5.68,3.936);
\foreach \y in {3.783,3.881} \draw[bcod] (\fovbx,\y-0.035) rectangle (5.60,\y+0.035);
\node[tag, inner sep=0pt, text=oipurple!75!black, anchor=west] at (0.74,3.832) {coda};
\node[ans, anchor=south] (outA) at (3.00,4.56) {A dog.};
\draw[fl] (3.00,\fovLt) -- (outA.south);

\def\fovL{6.50}\def\fovR{9.50}\def\fovcx{8.00}%
\def\fovco{0.13}%
\def\fovxj{10.27}%
\def\fovfb{1.55}%
\def\fovys{1.60}%
\colorlet{oiloop}{oistate!70!black}%
\tikzset{
  lwire/.style={draw=oiloop, line width=0.9pt, rounded corners=3pt},
  lhead/.style={-{Stealth[length=2.0mm,width=1.6mm]}},
}
\def\fovframe#1#2#3#4#5#6{%
  \draw[#1] (\fovL,#3) rectangle (\fovR,#4);
  \node[anchor=north west, inner sep=0pt, text height=2.0mm, text depth=0.6mm] at (\fovL+0.11,#4-0.03) {#5};
  \foreach \k in {1,...,#6}
    \draw[#2] (\fovL+0.08,{#4-0.30-0.098*(\k-1)-0.07}) rectangle (\fovR-0.08,{#4-0.30-0.098*(\k-1)});
}
\fovtokstrip{\fovcx-1.26}{0.35}
\draw[fl] (\fovcx,0.52) -- (\fovcx,0.76);
\fovframe{gpre}{bpre}{0.76}{1.243}{prelude}{2}
\draw[fill=oiverm!35, draw=oiverm!70!black, line width=0.6pt, rounded corners=1.2pt]
     (\fovcx-0.72,\fovfb) -- (\fovcx+0.72,\fovfb) -- (\fovcx+0.30,\fovfb+0.44)
     -- (\fovcx-0.30,\fovfb+0.44) -- cycle;
\draw[ybar] (\fovcx-0.67,\fovfb) rectangle (\fovcx-0.10,\fovfb+0.11);
\draw[sbar] (\fovcx+0.10,\fovfb) rectangle (\fovcx+0.67,\fovfb+0.11);
\draw[sbar] (\fovcx-0.18,\fovfb+0.44) rectangle (\fovcx+0.18,\fovfb+0.51);
\node[font=\footnotesize, anchor=east, inner sep=0pt] at (\fovcx-0.58,\fovfb+0.30) {adapter};
\draw[fl, draw=oiyellow!60!black] (\fovcx-0.29,1.243) -- (\fovcx-0.29,\fovfb);
\def\fovnx{8.29}\def\fovny{1.343}%
\draw[strip, line width=0.4pt, rounded corners=1.2pt] (\fovnx-0.35,\fovny-0.055) rectangle (\fovnx+0.35,\fovny+0.055);
\foreach \g [count=\i from -3] in {15,45,30,60,20,50,35}
  \fill[black!\g] ({\fovnx+0.095*\i-0.04},\fovny-0.04) rectangle ({\fovnx+0.095*\i+0.04},\fovny+0.04);
\draw[black!70, line width=0.6pt, dash pattern=on 0.6pt off 1.2pt, -{Stealth[length=1.6mm,width=1.3mm]}] (\fovnx,\fovny+0.055) -- (\fovnx,\fovfb);
\node[tag, text=black!70, anchor=south west] at (\fovnx+0.39,1.303) {initial state};
\draw[line width=0.65pt, draw=black!75] (\fovcx,\fovfb+0.51) -- (\fovcx,2.103);
\fovstatestrip{\fovcx}{2.158}
\def\fovbb{2.478}\def\fovbt{3.172}%
\draw[fl] (\fovcx,2.213) -- (\fovcx,\fovbb);
\foreach \k in {2,1}
  \draw[rounded corners=2pt, line width=0.7pt, fill=oigreen!14, draw=oigreen!75!black]
       (\fovL+\k*\fovco,\fovbb+\k*\fovco) rectangle (\fovR+\k*\fovco,\fovbt+\k*\fovco);
\fovframe{gblk}{bblk}{\fovbb}{\fovbt}{recurrent block}{4}
\draw[line width=0.65pt, draw=black!75] (\fovcx,\fovbt) -- (\fovcx,3.462);
\fovstatestrip{\fovcx}{3.517}
\def\fovjy{3.517}%
\draw[fl] (\fovcx,3.572) -- (\fovcx,3.837);
\fovframe{gcod}{bcod}{3.837}{4.32}{coda}{2}
\node[ans, anchor=south] (outB) at (\fovcx,4.56) {A dog.};
\draw[fl] (\fovcx,4.32) -- (outB.south);
\def\fovic{2.605}%
\draw[lwire, lhead] (\fovcx+0.47,\fovjy) -- (\fovxj,\fovjy) -- (\fovxj,\fovys) -- (\fovcx+0.67,\fovys);
\def\fovfl{10.75}\def\fovfr{13.65}\def\fovbb{0.35}\def\fovft{4.90}%
\draw[gdot, draw=black!35, fill=white, fill opacity=0.6, draw opacity=1, sharp corners]
     (\fovfl,\fovic-0.11) [rounded corners=2pt] -- (\fovfl,\fovbb) -- (\fovfr,\fovbb) -- (\fovfr,\fovft)
     -- (\fovfl,\fovft) -- (\fovfl,\fovic+0.11) [sharp corners] -- (\fovxj+0.40,\fovic) -- cycle;
\node[tag, text=black!60, anchor=north west, inner sep=0pt] at (\fovfl+0.11,\fovft-0.05) {unrolled};
\def\fovmx{12.30}%
\draw[strip, line width=0.35pt, rounded corners=1pt] (\fovmx-0.39,0.47) rectangle (\fovmx+0.33,0.59);
\foreach \q in {-3,...,0}
  \fill[oiblue, draw=oiblue!70!black, line width=0.25pt] (\fovmx+0.10*\q-0.04,0.49) rectangle (\fovmx+0.10*\q+0.04,0.57);
\foreach \q in {1,2,3}
  \draw[oiorange, line width=0.4pt] (\fovmx+0.10*\q-0.037,0.493) rectangle (\fovmx+0.10*\q+0.037,0.567);
\draw[mfl] (\fovmx,0.59) -- (\fovmx,0.77);
\draw[mpre] (\fovmx-0.95,0.77) rectangle (\fovmx+0.95,0.97);
\foreach \py in {0.04,-0.04}
  \draw[line width=0.25pt, fill=oiyellow!45, draw=oiyellow!60!black, rounded corners=0.4pt]
       (\fovmx-0.80,0.87+\py-0.02) rectangle (\fovmx+0.80,0.87+\py+0.02);
\draw[mybus] (\fovmx-0.70,0.97) -- (\fovmx-0.70,1.04) -- (\fovmx-1.12,1.04) -- (\fovmx-1.12,3.45);
\foreach \my in {1.17, 2.27, 3.42} {
  \draw[mybus, -{Stealth[length=1.2mm,width=1.0mm]}] (\fovmx-1.12,\my+0.03) -- (\fovmx-0.42,\my+0.03);
  \draw[mad] (\fovmx-0.46,\my+0.06) -- (\fovmx+0.46,\my+0.06) -- (\fovmx+0.20,\my+0.44) -- (\fovmx-0.20,\my+0.44) -- cycle;
  \draw[line width=0.3pt, fill=oiyellow!45, draw=oiyellow!60!black] (\fovmx-0.42,\my) rectangle (\fovmx-0.06,\my+0.06);
  \draw[line width=0.3pt, fill=oistate!45, draw=oistate!70!black] (\fovmx+0.06,\my) rectangle (\fovmx+0.42,\my+0.06);
  \draw[mfl] (\fovmx,\my+0.44) -- (\fovmx,\my+0.52);
  \draw[mblk] (\fovmx-0.95,\my+0.52) rectangle (\fovmx+0.95,\my+0.84);
  \foreach \py in {0.0975,0.0325,-0.0325,-0.0975}
    \draw[line width=0.25pt, fill=oigreen!45, draw=oigreen!60!black, rounded corners=0.4pt]
         (\fovmx-0.80,\my+0.68+\py-0.018) rectangle (\fovmx+0.80,\my+0.68+\py+0.018);
}
\draw[strip, line width=0.3pt, rounded corners=0.8pt] (\fovmx+0.56,1.14) rectangle (\fovmx+1.00,1.26);
\foreach \g [count=\i from 0] in {20,55,35,45}
  \fill[black!\g] (\fovmx+0.59+0.10*\i,1.16) rectangle (\fovmx+0.67+0.10*\i,1.24);
\draw[mdot] (\fovmx+0.56,1.20) -- (\fovmx+0.42,1.20);
\draw[mfl] (\fovmx+0.17,2.01) -- (\fovmx+0.17,2.27);
\foreach \y in {3.24,3.30,3.36} \fill[black!60] (\fovmx+0.17,\y) circle (0.022);
\draw[mfl] (\fovmx,4.26) -- (\fovmx,4.38);
\draw[mcod] (\fovmx-0.95,4.38) rectangle (\fovmx+0.95,4.58);
\foreach \py in {0.04,-0.04}
  \draw[line width=0.25pt, fill=oipurple!45, draw=oipurple!60!black, rounded corners=0.4pt]
       (\fovmx-0.80,4.48+\py-0.02) rectangle (\fovmx+0.80,4.48+\py+0.02);

\draw[black!45, densely dashed, line width=0.4pt] (5.75,\fovLt) -- (6.25,5.43);
\draw[black!45, densely dashed, line width=0.4pt] (5.75,\fovLb) -- (6.25,0.03);

\end{tikzpicture}

%% file: sections/02_related.tex
\vspace{-6pt}%
\section{Related Work}
\vspace{-5pt}%
\label{sec:related}

\paragraph{Depth recurrence and architectural retrofitting.}Depth recurrence reuses a block of layers multiple times, increasing the effective depth of the model without adding any additional parameters \citep{dehghani2018ut}. Previous attempts at creating recurrent LMs include Huginn \citep{geiping2025huginn} and Ouro \citep{ouro2025}. Huginn splits the model into three components: a \emph{prelude} that embeds the input into a latent state, a \emph{recurrent block} that iterates on that state, and a \emph{coda} that decodes the final state into output logits.  Ouro iterates the entire set of layers, and Loopie \citep{loopies2026} applies each layer recurrently instead of performing block-level looping. In all three cases, the model is trained from scratch without any pre-initialization by existing models. Two findings inform our design: a looped block benefits from a stable learned connector between successive passes \citep{parcae2026}, and an iterated layer contributes only a fraction of the capacity of a stored one \citep{schwethelm2026}, thus bounding what recurrence can recover from pruning. As an alternative to full-corpus pretraining, recurrence can instead be achieved by retrofitting a pretrained model. Within this regime, there exist two lines of work: (1) looping a middle block of the stack without pruning, for latent reasoning \citep{koishekenov2025etd}, and (2) looping after reducing the depth, either by keeping a block of the surviving layers \citep{bae2024relaxed,mcleish2025retrofitted} or by merging neighbouring layers into a few blocks that are then looped \citep{cherry2026}. We adopt the latter approach, retrofitting recurrence after structural pruning. Specifically, we take inspiration from \citet{mcleish2025retrofitted}, who drop layers 4 to 9 of TinyLlama-1.1B to obtain a $4/8/4$ layer allocation looped through a linear adapter.

\vspace{-2mm}
\paragraph{Recurrence in vision-language models.} 
Current works either loop layers of a pretrained VLM without pruning, with a near-constant number of parameters \citep{recursivevlm2026,reloopume2026}, or build one on a natively recurrent language backbone \citep{hive2026}.
HIVE \citep{hive2026} builds a VLM on the pretrained recurrent language model Huginn-3.5B \citep{geiping2025huginn} by attaching an InternViT-300M vision encoder and training the combined model on 6.5M training samples in three stages. 
As such HIVE is based on a significantly larger model and trained on five times more training samples than the proposed model, preventing a direct comparison to those checkpoints.
We therefore take inspiration from the pipeline in HIVE to replicate this setting as a baseline to compare to rather than comparing against their model. 
Beyond standard vision-language tasks, depth recurrence has been explored for robotic action policies, either in the action head of an intact VLM \citep{rdvla2026} or by pruning a VLM and looping a small block \citep{loopvla2026}. However, these methods operate in a fundamentally different domain, and the pruned variant of \citet{loopvla2026} lacks a prune-only baseline. %
Our setting departs from those existing approaches by following a prune-then-loop pattern, removing layers before the loop is installed. To the best of our knowledge, this work presents the first prune-then-loop retrofit of a pretrained VLM evaluated on VQA.

%% file: sections/03_method.tex
\vspace{-6pt}%
\section{Method: Prune-Then-Loop Retrofitting}\label{sec:method}
\vspace{-5pt}%

We study the problem of converting a pretrained non-recurrent vision-language model into a depth-recurrent architecture. In this section, we formalize the prune-then-loop retrofit, explaining the proposed model surgery, the training curriculum used to recover performance, and the adapter mechanism.

\vspace{-6pt}%
\subsection{Retrofitting existing VLMs}\label{sec:method:retrofit}
\vspace{-5pt}%

A pretrained vision-language model computes next-token logits from an image $I$ and a
text sequence $x$ as
\begin{equation}
F(I,x) \;=\; L(u),
\qquad L=\ell_{m}\circ\cdots\circ\ell_{1},
\qquad u=[\,\pi(V(I));\,E(x)\,],
\label{eq:vlm}
\end{equation}
where $V$ denotes the vision encoder, $\pi$ the projector, $E$ the text token embedder, and $L$ the language model composed of layers operating on representations of dimension $H$.
Our prune-then-loop mechanism modifies only the layers $\ell_{1},\dots,\ell_{m}$ of the language model $L$.
We partition the $m$ layers into four consecutive groups of sizes $p+d+b+c=m$ and delete the $d$ layers. This yields a \emph{prelude} block $P=\ell_{p}\circ\cdots\circ\ell_{1}$, a \emph{recurrent} block $B=\ell_{p+d+b}\circ\cdots\circ\ell_{p+d+1}$, and a \emph{coda} block
$C=\ell_{m}\circ\cdots\circ\ell_{m-c+1}$ (Figure~\ref{fig:overview}).

The retrofitted model applies the recurrent block $r$ times. The recurrent block is preceded by a small adapter $A$ that merges the block's current output with the output of the prelude. Expressing this pair as $B\circ A$ and repeating it $r$ times yields:
\begin{equation}
F^{\,r}_{p/b/c} \;\coloneqq\; C\circ(B\circ A)^{r}\circ P.
\label{eq:retrofit}
\end{equation}

Namely, we first apply the prelude $P$, then run the combination of adapter and recurrent block $B\circ A$ for $r$ times, and apply the coda $C$ once. We refer to $F^{\,r}_{p/b/c}$ as the retrofitted recurrent VLM (RR-VLM).
The prelude $P$ computes a context $h$ from the token input $u$ as $h = P(u)$. 
The adapter $A$ then concatenates the tokens of the recurrent block output $s$ with the tokens of prelude context $h$, resulting in the same number of tokens with double dimension, and then compresses the token-wise concatenated representation back to the original dimension $H$ through a learned linear layer $W\in\mathbb{R}^{H\times 2H}$ \citep{mcleish2025retrofitted}. In case of the first loop, when no recurrent output is computed yet, the tokens $s_{0}$ are initialized from a normal distribution $\mathcal{N}$ \citep{geiping2025huginn} and
$k$ is defined as the recurrence step: 
\begin{equation}
\label{eq:dynamics}
\begin{aligned}
A(s,h) &= W[s;h], && \text{linear adapter } A_{\mathrm{lin}} \\
s_{k} &= B\big(A(s_{k-1},h)\big), \quad s_{0}\sim\mathcal{N}(0,\sigma^{2}I),
  && \text{recurrent block, } k=1,\dots,r \\
y_{k} &= C(s_{k}), && \text{readout after } k \text{ recurrences}
\end{aligned}
\end{equation}
The coda $C$ maps the state after $k$ recurrences to output logits $y_{k}$, and the output of the model is $y_{r}$.
Note that every input token traverses $m^{r}=p+rb+c$ layers, but the model stores only $p+b+c$ distinct layers. Restricting this traversal to $r{=}1$ defines the single-pass model $F^{1}$. 

\paragraph{Number of recurrences.} The number of recurrences $r$ can be chosen flexibly during training and inference. During training, it can be defined in two ways: (1) as a constant, defining the model $F^{r}$, or (2) as a per-step sample from a log-normal Poisson distribution \citep{geiping2025huginn} with mean $\bar{r}$, defining the model $F^{\tilde r}$.
At inference, $r$ is set to a fixed value that can be independent of the training configuration.

\paragraph{KV cache.} Each pass through the recurrent block computes its own keys and
values, so the KV cache holds $m^{r}$ entries per token compared to the $m$ entries for the original
model. We evaluate two ways of sharing the cache across passes. The first, from
\citet{geiping2025huginn}, %
keeps $n$ copies of the block's cache per token and lets pass
$i$ use copy $i \bmod n$. With $n{=}1$, every pass reads the entries that the last pass
wrote, and the cache shrinks to $p+nb+c$ entries per token. This variant changes only the
inference procedure and requires no training. The second lets every pass reuse the keys,
or the keys and values, that the first pass computed, so that the cache shrinks to
$p+b+c$ entries per token. This variant changes the forward pass itself. The model is
therefore trained with the sharing active and evaluated with the same forward pass.

\vspace{-6pt}%
\subsection{Training Strategy}\label{sec:method:recovery}
\vspace{-5pt}%

The retrofitted model of Sec.~\ref{sec:method:retrofit} requires training for two reasons. First, the pruning removes the layers between the prelude and the recurrent block, leading to a misalignment between the representations produced by the prelude and those expected by the recurrent block. Additionally, the remaining layers have to be adapted to the modified architecture.
Second, the recurrence introduces an adapter $A$ that fuses the output of the prelude with the output of the recurrent block using a newly initialized linear layer that needs to be trained as well.

\vspace{-2mm}
\paragraph{Three-stage Training.} We propose a three-stage curriculum that restores the VLM and trains the adapter: First, in stage $S_{0}$ we focus on repairing by training only the pruned language backbone on text-only data, freezing all other components. Second, as the retrained text backbone can now slightly deviate from the original one, we retrain the vision-language projector in stage $S_{1}$ to restore image-text alignment, and finally optimize the entire model via end-to-end training with visual instruction data in stage $S_{2}$, interleaving image-text and text-only training data. 

\vspace{-2mm}
\paragraph{Knowledge Distillation.}
Retraining the modified language backbone inevitably shifts its internal representations away from those of the pretrained model. In a VLM this shift is problematic, since the vision-language projector has been aligned to the representations of the original backbone. We therefore seek to adapt the language backbone to the new architecture while preserving its original representations as far as possible. To this end, we incorporate a distillation term at stage $S_{0}$ training, using the original, unpruned baseline model $F$ as the teacher to limit the drift of the pruned, recurrent model $F^r$ and to allow an easier realignment in stage $S_{1}$:
\begin{equation}
\mathcal{L} \;=\; (1-\lambda)\,\mathcal{L}_{\mathrm{CE}}
\;+\; \lambda\,\tau^{2}\,
\operatorname{KL}\!\big(\operatorname{softmax}(F/\tau)\,\big\|\,\operatorname{softmax}(F^{r}/\tau)\big).
\label{eq:kd}
\end{equation}
The distillation objective evaluates a single pass of the teacher against the iterative execution of the student. Note that in stage $S_{0}$, this is done on text-only training data. %
We found that such knowledge distillation can further improve performance during stage $S_{2}$ training, as this combines the vision and language input signal from the teacher.  

%% file: sections/04_setup.tex
\vspace{-6pt}%
\section{Implementation Details}\label{sec:setup}
\vspace{-5pt}%

\phantomsection\label{sec:setup:training}%
\paragraph{Base model.}
We apply the retrofit approach to the pretrained
\texttt{bczhou/tiny-llava-v1-hf} checkpoint \citep{tinyllava2024}, named TinyLLaVA-1.4B by its authors, which comprises a
22-layer TinyLlama-1.1B language model \citep{tinyllama2024}, a CLIP ViT-L/14-336
vision tower \citep{clip2021}, and a LLaVA-style projector \citep{llava2023}. Note that \citet{mcleish2025retrofitted} retrofit the same language model in the text-only
setting into a recurrent language model. This allows us to use this as a base model for a directly comparable baseline of a recurrent language model that is extended by a vision backbone.

\phantomsection\label{sec:setup:surgery}%
\paragraph{Model Surgery.}
We again follow \citet{mcleish2025retrofitted} and delete six contiguous layers 4-9 ($d{=}6$) from the pretrained TinyLLaVA-1.4B checkpoint, leaving a 16-layer language model
with the layer allocation $p/b/c = 4/8/4$. The final model uses a fixed recurrence of $r{=}6$, resulting in each token traversing an equivalent depth of $m^{r}=4+6\times 8+4=56$ layers.
After surgery, the language model including the adapter contains 844.4M parameters compared to the 1,100.3M of the language backbone of the base model ($-23.3\%$). The adapter is a single linear layer with 8.4M parameters. Including the vision encoder and the projector the model has 1,154.2M parameters compared to 1,410.1M of the base model ($-18.1\%$).

\phantomsection\label{sec:setup:recovery}%
\paragraph{Training.}
The training follows the three-stage curriculum of Sec.~\ref{sec:method:recovery}. All three
stages use Muon \citep{muon2024} on matrix parameters and AdamW
\citep{loshchilov2019adamw} on the remainder.
The text-only training phase $S_{0}$ uses a 60B-token
blend of 85\% SlimPajama \citep{slimpajama2023} and 15\% FineWeb-Edu
\citep{penedo2024fineweb} for the main experiments, and on 7.5B tokens of SlimPajama for our ablations. The text-only training phase adds the KD term of Eq.~\ref{eq:kd} with $\lambda{=}0.5$ at $\tau{=}2$.
The image-text alignment phase $S_{1}$ trains on the base model's alignment set of
\texttt{blip\_sbu\_cc} \citep{blip2022} containing 493k image-caption pairs. It updates
only the projector $\pi$ and adds no KD term.
The visual instruction tuning phase $S_{2}$ finetunes the entire model on the base
model's instruction data of \texttt{llava\_instruct} \citep{llava2023} which consists of
660k cleaned samples, interleaving four multimodal batches and one text batch (Appendix~\ref{app:interleave} varies this ratio). It applies
the KD term again, with a separate weight and temperature for each batch type:
$\lambda{=}0.25$ at $\tau{=}2$ on multimodal batches and $\lambda{=}0.5$ at $\tau{=}1$ on text batches.
All models are trained with seed 42.

%% file: sections/05_results.tex
\vspace{-6pt}%
\section{Evaluation}\label{sec:eval}
\vspace{-5pt}%

\paragraph{Benchmarks.}\label{sec:setup:metric}
We evaluate the models on the following nine benchmarks: GQA \citep{gqa2019}, POPE \citep{pope2023}, TextVQA (TVQA) \citep{textvqa2019}, and ScienceQA-IMG (SQA) \citep{scienceqa2022}, MMBench (MMB) \citep{mmbench2023}, RealWorldQA (RWQA) \citep{realworldqa2024}, SEED-Bench (SEED) \citep{seedbench2023}, AI2D \citep{ai2d2016}, and the 2D split of CV-Bench (CV2D) \citep{tong2024cambrian}. 
We use the VLMEvalKit \citep{vlmevalkit2024} library for all model evaluations. 
We define the Avg metric as the arithmetic mean of the nine VQA benchmarks. 
We set a fixed eval seed to make evaluation deterministic.

\vspace{-6pt}%
\subsection{Comparison to Baselines}\label{sec:results:headline}
\vspace{-5pt}%

\input{tables/tab_main}
\phantomsection\label{sec:setup:baselines}%
\paragraph{Baselines.}
We evaluate the proposed recurrence retrofitting approach against five baselines and report the results in Table~\ref{tab:main}. The first baseline is the pretrained TinyLLaVA-1.4B checkpoint without any modifications. The second and third baselines fine-tune this model without pruning, either with the image-text stages $S_1$ and $S_2$ only or with all three stages $S_0$, $S_1$ and $S_2$ of our training strategy. The fourth baseline applies all three stages to the pruned model without recurrence, which corresponds to the single-pass model $F^{1}$ of Sec.~\ref{sec:method:retrofit}. The baselines with the text-only stage apply the distillation term of Eq.~\ref{eq:kd} in $S_0$, and the unpruned Baseline+$(S_0,S_1,S_2)$ also applies it in $S_2$, so that it is trained with the full strategy of our model. We include those retrained versions to allow a direct comparison between the retrofitted recurrent model and the same non-recurrent backbones trained on the same data.
To allow a comparison to a setup that starts from a recurrent language model and adds the vision backbone after recurrent training, we convert a pretrained recurrent TinyLlama checkpoint of \citet{mcleish2025retrofitted} into a recurrent VLM by adding the pretrained image encoder from
TinyLLaVA-1.4B and train the model with the same image-text data on the training stages $S_{1}$ and $S_{2}$.
For our retrofitted recurrent VLM (RR-VLM), we use a fixed recurrence of $r{=}6$ and report results for %
an 8-layer recurrent block architecture ($F^{6}_{4/8/4}$).

We see that the training improves the baseline models by around $2.4$ to $2.9$ pp, while the performance of the pruned model is slightly below that of the base model. 
The performance of the converted recurrent VLM  is in the same range as the retrained baselines. 
Compared to that, the proposed RR-VLM model is able to outperform all baseline models on average by at least $2.3$ pp. 
Analyzing the performance per dataset, we see that more basic visual understanding datasets, such as GQA, POPE, TVQA and SQA, do not profit as much from the recurrent architecture, and that the RR-VLM performs on par with the baseline model. This can be attributed to the fact that those datasets usually have single-step QA type questions, which do not profit from more complex reasoning. Recurrence adds computation without adding information, so no gain is expected on such questions. 

Compared to that, we see a clear trend of performance increase on more reasoning-oriented datasets such as MMBench, RealWorldQA, SEED-Bench, AI2D and CV2D. Note that on the latter four datasets, both recurrent models, the converted VLM as well as the proposed RR-VLM model, perform consistently better than the non-recurrent baselines, with RR-VLM still outperforming the post hoc converted VLM on three of the four datasets. We thus find that retrofitting a pretrained VLM results in better performance than introducing a vision encoder into a recurrent language model to create a recurrent VLM.  
We thus find that retrofitting a pretrained VLM can result in improved performance compared to introducing a vision encoder into a recurrent language model to create a recurrent VLM.

\vspace{-6pt}%
\subsection{Ablations}\label{sec:results:ablations}

\input{tables/tab_stages}
\paragraph{Training stages.}

We determine the effect of the text-only stage $S_0$ and the vision-language alignment stage $S_1$ on the resulting recurrent VLM in Table~\ref{tab:stages}. We find that $S_0$ is crucial to the recovery training, as removing it reduces the performance of the resulting model by $3.68$ pp on average. Removing $S_0$ primarily affects the reasoning-oriented datasets, with $7.3$ pp lost on CV2D, $6.8$ pp on MMB and $4.6$ pp on AI2D. We attribute this to the fact that without the repair of the pruned language backbone in $S_0$, the stages $S_1$ and $S_2$ have to repair the language backbone and align the projector at the same time. 
We further find that removing $S_1$ costs $1.28$ pp, again primarily on the reasoning-oriented datasets. We attribute this to the fact that $S_1$ performs the vision-language alignment, and without it the stage $S_2$ has to perform both the alignment and the instruction tuning. 
In summary, both $S_0$ and $S_1$ contribute to the performance, with $S_0$ having the larger impact, which shows that the text-only recovery is necessary for retrofitting recurrence into a pruned VLM.

\vspace{-2mm}
\input{tables/tab_ablation}

\paragraph{Layer allocation.} We first ablate the design of the recurrent block in Table~\ref{tab:ablation}(a). To this end, we fix the depth of the recurrent model at $m^{r}{=}56$ layers, thus smaller recurrent blocks are executed with more recurrences compared to larger ones. All models are trained on a $7.5B$ $S_0$ text-token budget. We find that a larger recurrent block performs better than other variants given the same compute setting. Namely, the $4/8/4$ allocation outperforms a $6/4/6$
allocation at $r{=}11$ by $3.19$ pp. A narrower $4/6/4$ block at $r{=}8$ prunes two additional layers, and the $4/8/4$ configuration outperforms it by $1.60$ pp.

\vspace{-2mm}
\paragraph{Adapters.}
Next, we consider the adapter design in Table~\ref{tab:ablation}(b), comparing the proposed linear adapter of Eq.~\ref{eq:dynamics} against multiple adapter variants, including the case of no adapter. 
The gated adapter $A_{\mathrm{sc}}$ adds the context $h$ to the state $s$, multiplied channel-wise by a learned factor $(1+\alpha)$ with $\alpha$ initialized at zero. Thus the adapter starts as a plain sum: $A_{\mathrm{sc}}(s,h)=s+(1+\alpha)\odot h$. 
The Parcae adapter $A_{\mathrm{pc}}$ follows the adapter design of \citet{parcae2026}. It multiplies the state by a learned per-channel factor $a\in(0,1)^{H}$, so that the state of the previous recurrence is damped and cannot grow across recurrences, and adds a learned projection $W'$ of the normalized context, weighted by a learned per-channel scale $\delta$: $A_{\mathrm{pc}}(s,h)=a\odot s+\delta\odot W'\,\mathrm{LN}(h)$. 
The no adapter variant $A_{\varnothing}$ starts the loop from $s_{0}=h$ instead of noise. The evaluation is done on a 4/6/4 architecture with $r=6$ and all models are trained on a $7.5B$ $S_0$ text-token budget. 
Overall, it shows that the linear adapter $A_{\mathrm{lin}}$ (Eq.~\ref{eq:dynamics}) achieves the highest performance. 
The adapter-free configuration $A_{\varnothing}$ initializes the loop directly from the prelude output $h$ and noticeably performs nearly as well as $A_{\mathrm{lin}}$ with a performance gap of only $0.73$ pp. 
Furthermore, $A_{\varnothing}$ outperforms both the gated adapter $A_{\mathrm{sc}}$ and the Parcae adapter $A_{\mathrm{pc}}$ by $1.31$ pp and $0.53$ pp, respectively. 

\vspace{-2mm}
\paragraph{Looping Variants.} 
We compare block-wise looping with per-layer looping in Table~\ref{tab:ablation}(c). Per-layer looping loops each layer of the recurrent block individually \citep{loopies2026} instead of the block as a whole. Namely, we again consider a 4/6/4 architecture trained on a $7.5B$ $S_0$ text-token budget, but with per-layer looping for $r{=}2$ and $r{=}6$ recurrences.
Overall, it shows that block looping outperforms per-layer looping by $2.83$ pp at $r{=}2$ and $5.40$ pp at $r{=}6$ and that per-layer looping does not profit from an increase of recurrences, with the performance dropping with more loops.

\input{tables/tab_kd}

\vspace{-2mm}
\paragraph{Knowledge distillation.} 
We further evaluated the impact of knowledge distillation, especially at the training stage $S_2$, and compared this setting with no knowledge distillation, a knowledge distillation based on image-text samples only, as well as a knowledge distillation with text-image as well as text-only training data in Table~\ref{tab:kd}.  
Overall, it shows that knowledge distillation in $S_2$ always improves the overall performance, and that having both image-text as well as text-only data leads to the most significant improvement, improving over the undistilled version by $1.48$ pp. 
The unpruned baseline gains only $0.25$ pp from the same $S_2$ distillation (first two rows). We attribute the gain of the recurrent model to the fact that the knowledge distillation acts as a form of regularizer for the new recurrent architecture, which keeps the model close to the pretrained one and prevents it from overfitting on the image-text data. This is in line with the large effect of removing $S_0$ in Table~\ref{tab:stages}, which also anchors the retrofitted model to the pretrained one. Appendix~\ref{app:kd75} reports the placement of the distillation term across the training stages at the 7.5B budget.

\input{tables/tab_depth}

\vspace{-2mm}
\phantomsection\label{sec:results:depth}\label{par:depth}%
\paragraph{Fixed versus sampled depth.}
We further evaluate the effect of the number of recurrences during training in Table~\ref{tab:depth}. To this end, we compare a model trained with a fixed number of recurrences $r{=}6$ against a model trained with a variable number of recurrences, sampled per training step with mean $6$ (Sec.~\ref{sec:method:retrofit}). Both models are evaluated at test-time depths from $r{=}2$ to $r{=}16$.
We find that the model trained with a fixed recurrence only works at the depth it was trained for, degrades at the neighbouring depths and collapses beyond $r{=}8$ or below $r{=}5$. Its GQA accuracy drops from $59.7$ at $r{=}6$ to $43.1$ at $r{=}5$ and $50.9$ at $r{=}8$, and the output degenerates at all other depths. In contrast, the model trained with a sampled recurrence regime also works at depths other than the one it was trained for, staying within $2$ pp of its best value from $r{=}2$ to $r{=}16$. This can be attributed to the fact that the sampled regime exposes the model to a different number of recurrences in every training step, so that its readout is trained for every depth in that range. However, we find that its performance saturates and does not improve beyond the trained depth. Consequently, we do not observe test-time scaling. %
The fixed and sampled recurrence models perform on par at the trained depth, with $55.75$ against $55.83$ on Avg. We therefore find that a fixed recurrence model is the less expensive approach to train a recurrent VLM, but it comes at the price of test-time stability to varying recurrences. Appendix~\ref{app:ttc} evaluates early exit for the sampled model.

\input{tables/tab_kvzero}
\phantomsection\label{sec:results:costs}\label{sec:results:cost}\label{sec:results:knowledge}%
\paragraph{KV-Cache Sharing.}
Finally, we evaluate the effect of sharing the KV cache across recurrences at inference in Table~\ref{tab:kvzero}. To this end, we apply the inference-time sharing of Sec.~\ref{sec:method:retrofit} to our trained model and reduce the number of cache copies $n$ from one per recurrence, which is our setting, down to a single copy. Since every pass through the recurrent block keeps its own keys and values, our model holds 2.55$\times$ the KV cache of the base model, in line with the 56 against 22 traversed layers per token, while it stores 18.1\% fewer weights and outperforms it (Table~\ref{tab:main}).
We see that the cost grows as the budget $n$ shrinks. With $n{=}3$ the cache shrinks to 1.45$\times$ that of the base model at a cost of $0.75$ pp, and with $n{=}1$ it shrinks to 0.727$\times$, below the base model itself, at a cost of $5.43$ pp. This can be attributed to the fact that the keys and values of a recurrence step are computed from the state of that step, so a recurrence step that reads the cache of another step attends to representations that do not match its own input. Training the model with the sharing active does not close this gap either, with a cost of $0.73$ pp for sharing the keys and $2.94$ pp for sharing the keys and values (Appendix~\ref{app:kvtrain}). We thus keep a separate cache per recurrence in our best-performing setting, and note that the cache size can be traded against accuracy at inference without retraining. Table~\ref{tab:kvtrain} in the appendix reports sharing applied during training.

\vspace{-6pt}%
\subsection{Qualitative examples}\label{sec:results:qual}
\vspace{-5pt}%
Figure~\ref{fig:qual} shows one item from each of the four reasoning-oriented benchmarks on which our model answers correctly and all three trained non-recurrent baselines of Table~\ref{tab:main} answer wrongly. We compare our model to the pretrained model and to the converted recurrent VLM, which is our closest recurrent baseline. For further qualitative examples, we refer to Appendix~\ref{app:qual}.
\input{tables/fig_qual_row_c}

%% file: tables/tab_main.tex
\begin{table}[tb]
\caption{\textbf{Comparison to baselines.} The retrofit outperforms the baselines on average over nine benchmarks with 23.3\% fewer stored language-model parameters.  Sec.~\ref{sec:results:headline} defines the models. RR-VLM denotes our retrofitted recurrent VLM $F^{6}_{4/8/4}$ of Sec.~\ref{sec:method} and $F^{1}$ the pruned single-pass model.
}
\label{tab:main}
\centering
\scriptsize
\setlength{\tabcolsep}{1.7pt}%
\begin{tabular*}{\textwidth}{@{\extracolsep{\fill}}l cccccccccc|c@{}}
\toprule
\textbf{\boldmath Model} & \textbf{\boldmath LM (B)} & \textbf{\boldmath GQA} & \textbf{\boldmath POPE} & \textbf{\boldmath TVQA} & \textbf{\boldmath SQA} & \textbf{\boldmath MMB} & \textbf{\boldmath RWQA} & \textbf{\boldmath SEED} & \textbf{\boldmath AI2D} & \textbf{\boldmath CV2D} & \textbf{\boldmath Avg} \\
\midrule
\multicolumn{12}{@{}l}{\emph{Baseline}} \\
\addlinespace[1pt]
{TinyLLaVA-1.4B} & 1.10 & 57.6 & 85.1 & 46.7 & 59.6 & 49.7 & 45.4 & 54.1 & 39.4 & 38.6 & 52.91 \\
\midrule
\multicolumn{12}{@{}l}{\emph{Baselines, no recurrence}} \\
\addlinespace[1pt]
{Baseline+$(S_1,S_2)$} & 1.10 & 59.9 & \textbf{86.9} & 45.5 & 57.8 & 55.6 & 47.3 & 59.8 & 41.3 & 43.4 & 55.27 \\
{Baseline+$(S_0,S_1,S_2)$} & 1.10 & \textbf{60.4} & 86.7 & 47.3 & \textbf{60.2} & 55.5 & 46.4 & 60.4 & 43.2 & 42.2 & 55.80 \\
{Baseline(pruned)+$(S_0,S_1,S_2)$ ($F^{1}$)} & 0.84 & 58.0 & 86.1 & 37.7 & 50.3 & 48.6 & 47.6 & 58.3 & 38.1 & 47.1 & 52.41 \\
\midrule
\multicolumn{12}{@{}l}{\emph{Converted recurrent LM \citep{mcleish2025retrofitted}}} \\
\addlinespace[1pt]
{Recurrent LM+ViT+$(S_1,S_2)$} & 0.84 & 58.8 & 86.3 & 41.5 & 52.5 & 52.7 & \textbf{50.6} & 60.7 & 43.6 & 53.7 & 55.59 \\
\midrule
\multicolumn{12}{@{}l}{\emph{Pruned and looped (ours)}} \\
\addlinespace[1pt]
\rowcolor{gray!12}{RR-VLM $F^{6}_{4/8/4}$} & 0.84 & 59.2 & 86.7 & \textbf{47.7} & 59.1 & \textbf{58.9} & 50.5 & \textbf{62.6} & \textbf{44.6} & \textbf{53.9} & \textbf{58.13} \\
\bottomrule
\end{tabular*}
\end{table}

%% file: tables/tab_stages.tex
\begin{table}[t]
\caption{\textbf{Training stages.} We remove single stages of the training strategy of Sec.~\ref{sec:method:recovery} from our best performing setting ($4/8/4$, $r{=}6$, 60B $S_0$ budget), each stage with its distillation term. The first row repeats the unpruned baseline of Table~\ref{tab:main}, which is trained with the full strategy. Avg is the mean of the nine columns to its left.}
\label{tab:stages}
\centering
\scriptsize
\setlength{\tabcolsep}{2pt}
\begin{tabular*}{\textwidth}{@{\extracolsep{\fill}}l cccccccccc@{}}
\toprule
\textbf{\boldmath Model and training stages} & \textbf{\boldmath GQA} & \textbf{\boldmath POPE} & \textbf{\boldmath TVQA} & \textbf{\boldmath SQA} & \textbf{\boldmath MMB} & \textbf{\boldmath RWQA} & \textbf{\boldmath SEED} & \textbf{\boldmath AI2D} & \textbf{\boldmath CV2D} & \textbf{\boldmath Avg} \\
\midrule
Baseline+$(S_0,S_1,S_2)$ & 60.4 & 86.7 & 47.3 & 60.2 & 55.5 & 46.4 & 60.4 & 43.2 & 42.2 & 55.80 \\
\midrule
RR-VLM, $S_1$, $S_2$ & 59.6 & 86.5 & 45.5 & 54.6 & 52.1 & 44.1 & 61.0 & 40.0 & 46.6 & 54.45 \\
RR-VLM, $S_0$, $S_2$ & 59.3 & 86.6 & 47.0 & 57.1 & 56.7 & 48.8 & 62.4 & 42.7 & 51.2 & 56.85 \\
\rowcolor{gray!12}RR-VLM, $S_0$, $S_1$, $S_2$ (ours) & 59.2 & 86.7 & 47.7 & 59.1 & 58.9 & 50.5 & 62.6 & 44.6 & 53.9 & \textbf{58.13} \\
\bottomrule
\end{tabular*}
\end{table}

%% file: tables/tab_ablation.tex
\begin{table}[t]
\caption{\textbf{Ablations.} Each table changes one setting of the model named in its header and the best value is in bold. All tables (a), (b) and (c) use the 7.5B $S_{0}$ budget without text distillation.}
\label{tab:ablation}
\centering
\footnotesize
\newcommand{\ablock}[1]{\multicolumn{2}{@{}l}{\textbf{#1}} \\ \midrule}
\begin{minipage}[t]{0.32\textwidth}
\begin{tabular*}{\linewidth}[t]{@{\extracolsep{\fill}}lr@{}}
\toprule
\ablock{(a) Layer allocation, 56 layers}
$6/4/6$, $r{=}11$ & 52.56 \\
$4/6/4$, $r{=}8$ & 54.15 \\
$4/8/4$, $r{=}6$ & \textbf{55.75} \\
\bottomrule
\end{tabular*}
\end{minipage}
\hfill
\begin{minipage}[t]{0.32\textwidth}
\begin{tabular*}{\linewidth}[t]{@{\extracolsep{\fill}}lr@{}}
\toprule
\ablock{(b) Adapter, $4/6/4$, $r{=}6$}
$A_{\varnothing}$ (none) & 54.29 \\
$A_{\mathrm{sc}}$ & 52.98 \\
$A_{\mathrm{pc}}$ & 53.76 \\
$A_{\mathrm{lin}}$ & \textbf{55.02} \\
\bottomrule
\end{tabular*}
\end{minipage}
\hfill
\begin{minipage}[t]{0.32\textwidth}
\begin{tabular*}{\linewidth}[t]{@{\extracolsep{\fill}}lr@{}}
\toprule
\ablock{(c) Looping variant, $4/6/4$}
per-layer looping, $r{=}2$ & 52.19 \\
per-layer looping, $r{=}6$ & 49.62 \\
block looping, $r{=}6$ & \textbf{55.02} \\
\bottomrule
\end{tabular*}
\end{minipage}
\end{table}

%% file: tables/tab_kd.tex
\begin{table}[t]
\caption{\textbf{Knowledge distillation.} We evaluate the impact of the knowledge distillation in $S_2$. Each row states on which batches the distillation term is applied and with which weight $\lambda$ and temperature $\tau$. All rows use the 60B $S_0$ budget and the $4/8/4$ layer allocation and share the same $S_0$, which applies the distillation term with $\lambda{=}0.5$ at $\tau{=}2$. The first two rows give the unpruned baseline of Table~\ref{tab:main} with and without the $S_2$ distillation.}
\label{tab:kd}
\centering
\footnotesize
\begin{tabular*}{0.6\textwidth}{@{\extracolsep{\fill}}lr@{}}
\toprule
\textbf{Knowledge distillation} & \textbf{Avg} \\
\midrule
Baseline+$(S_0,S_1,S_2)$ (Table~\ref{tab:main}) & 55.80 \\
\quad without $S_2$ distillation & 55.55 \\
\midrule
RR-VLM, no $S_2$ distillation & 56.65 \\
RR-VLM, VL-KD only, $\lambda{=}0.25$ & 56.88 \\
\quad$+$ text-KD, $\tau{=}2$ & 57.24 \\
\quad$+$ text-KD, $\tau{=}1$ (ours) & \textbf{58.13} \\
\bottomrule
\end{tabular*}
\end{table}

%% file: tables/tab_depth.tex
\begin{table}[t]
\caption{\textbf{Fixed against sampled training depth.} Both models use the $4/8/4$ layer allocation and the 7.5B $S_0$ budget and differ only in the number of recurrences during training, fixed at $r{=}6$ or sampled with mean $6$. The columns give GQA accuracy when the model is evaluated with $r$ passes. Values marked $^{\dagger}$ come from runs that were stopped early because the output had degenerated. %
}
\label{tab:depth}
\centering
\footnotesize
\setlength{\tabcolsep}{4pt}
\begin{tabular}{@{}lrrrrrrr@{}}
\toprule
& \multicolumn{7}{c}{\textbf{\boldmath GQA at test-time depth $r$}} \\
\cmidrule(l){2-8}
\textbf{\boldmath Training depth} &  \textbf{\boldmath $r{=}2$} & \textbf{\boldmath $r{=}4$} & \textbf{\boldmath $r{=}5$} & \textbf{\boldmath $r{=}6$} & \textbf{\boldmath $r{=}8$} & \textbf{\boldmath $r{=}12$} & \textbf{\boldmath $r{=}16$} \\
\midrule
sampled, $\bar{r}{=}6$ ($F^{\tilde r}$) &  57.8 & 59.3 & 59.5 & 59.5 & 59.4 & 59.1 & 59.3 \\
\rowcolor{gray!12}fixed, $r{=}6$ ($F^{r}$, our setting) &  0.0$^{\dagger}$ & 0.0$^{\dagger}$ & 43.1 & 59.7 & 50.9 & 4.2$^{\dagger}$ & 0.0$^{\dagger}$ \\
\bottomrule
\end{tabular}
\end{table}

%% file: tables/tab_kvzero.tex
\begin{table}[t]
\caption{\textbf{KV-cache sharing across passes at inference.} The table analyses the KV
cache per token as a multiple of the base model's cache. The model
keeps only $n$ copies of the cache instead of one per pass. The six passes use these copies
in turn, so with $n{=}1$ every pass shares one cache, and
$n{=}6$ is our model without sharing. }
\label{tab:kvzero}
\centering
\scriptsize
\renewcommand{\arraystretch}{1.0}
\setlength{\tabcolsep}{0.5pt}
\begin{tabular*}{\textwidth}{@{\extracolsep{\fill}}@{}lrw{c}{3.1em}w{c}{3.1em}w{c}{3.1em}w{c}{3.1em}w{c}{3.1em}w{c}{3.1em}w{c}{3.1em}w{c}{3.1em}w{c}{3.1em}r@{}}
\toprule
\textbf{\boldmath Setting} & \textbf{\boldmath Cache} & \textbf{\boldmath GQA} & \textbf{\boldmath POPE} & \textbf{\boldmath TVQA} & \textbf{\boldmath SQA} & \textbf{\boldmath MMB} & \textbf{\boldmath RWQA} & \textbf{\boldmath SEED} & \textbf{\boldmath AI2D} & \textbf{\boldmath CV2D} & \textbf{\boldmath Avg} \\
\midrule
$n{=}1$ (last pass) & $0.727\times$ & 49.1 & 86.7 & 18.6 & 59.3 & 57.0 & 47.7 & 61.2 & 42.2 & 52.5 & 52.70 \\
$n{=}2$ & $1.09\times$ & 56.5 & 86.7 & 31.5 & 59.2 & 58.6 & 49.7 & 62.6 & 44.1 & 53.8 & 55.85\\
$n{=}3$ & $1.45\times$ & 58.6 & 86.7 & 41.8 & 59.4 & 58.6 & 50.5 & 62.6 & 44.4 & 53.9 & 57.38 \\
\rlap{\color{gray!11}\rule[\dimexpr-\arraystretch\dp\strutbox\relax]{\textwidth}{\dimexpr\arraystretch\ht\strutbox+\arraystretch\dp\strutbox\relax}}$n{=}6$ (ours) & $2.55\times$ & 59.2 & 86.7 & 47.7 & 59.1 & 58.9 & 50.5 & 62.6 & 44.6 & 53.9 & 58.13 \\
\bottomrule
\end{tabular*}
\end{table}

%% file: tables/fig_qual_row_c.tex
\input{tables/qual_macros}

\providecommand{\qshort}[6]{%
\setlength{\tabcolsep}{0pt}%
\begin{tabular}[H]{@{}l@{\hspace{3pt}}l@{}}
TinyLLaVA-1.4B & #1 \\
Recurrent LM+ViT & #5 \\
\rowcolor{gray!12}RR-VLM (ours) & #6 \\
\end{tabular}}
\providecommand{\qitem}[5]{%
\begin{minipage}[t]{0.235\textwidth}
\vspace{0pt}
\includegraphics[width=\linewidth]{#1}\par
\begin{minipage}[t]{\linewidth}\raggedright
\textbf{#2}\\ #3 #4
\end{minipage}\par\vspace{2pt}
#5
\end{minipage}}
\begin{figure}[H]
\centering
\scriptsize
\qitem{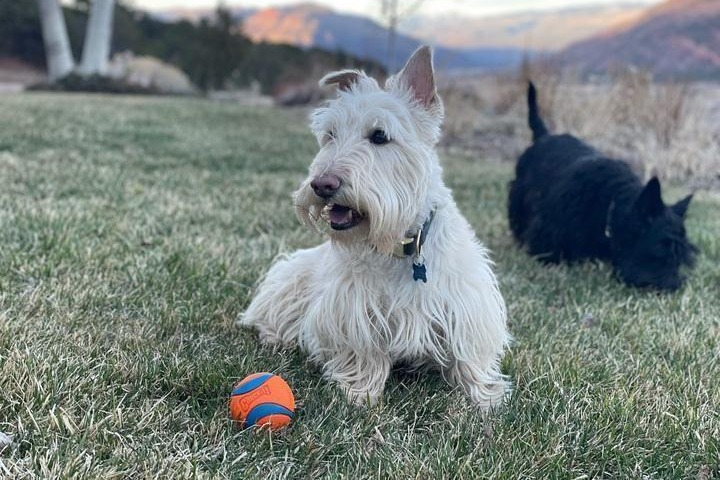}{RealWorldQA}
{Where is the orange ball located relative to the white dog?}
{A.~left of the dog \quad B.~right of the dog \quad C.~behind the dog}
{\qshort{\qno{C}}{\qno{C}}{\qno{C}}{\qno{C}}{\qok{A}}{\qok{A}}}\hfill
\qitem{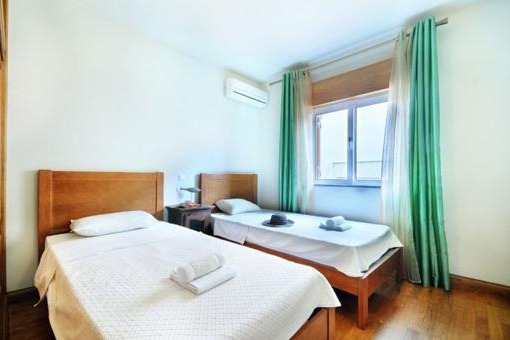}{SEED-Bench}
{How many pillows are on the bed closest to the window?}
{A.~0 \quad B.~1 \quad C.~3 \quad D.~2}
{\qshort{\qno{D}}{\qno{D}}{\qno{D}}{\qno{D}}{\qno{D}}{\qok{B}}}\hfill
\qitem{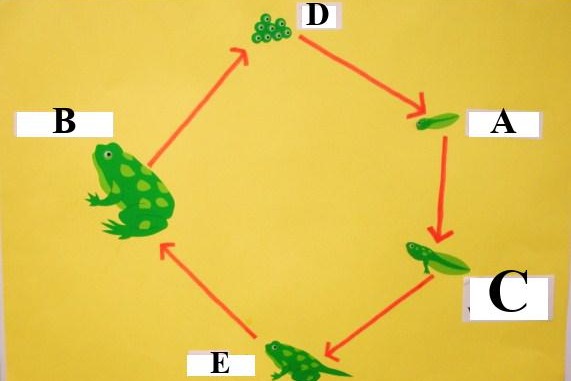}{AI2D}
{What stage is between a tadpole with legs and an adult frog?}
{A.~adult frog \quad B.~tadpole \quad C.~egg \quad D.~froglet}
{\qshort{\qno{B}}{\qno{C}}{\qno{A}}{\qno{B}}{\qno{B}}{\qok{D}}}\hfill
\qitem{}{CV-Bench-2D}
{Considering the relative positions of the flowerpot and the picture, where is the flowerpot located with respect to the picture?} {A.~left \quad B.~right}
{\qshort{\qok{B}}{\qno{A}}{\qno{A}}{\qno{A}}{\qno{A}}{\qok{B}}}
\caption{\textbf{Qualitative examples.} One item from each of the four reasoning-oriented benchmarks %
with the answers of the pretrained model, the converted recurrent model and our model below the image, a check mark denotes the correct option. The three trained non-recurrent baselines of Table~\ref{tab:main} answer all four items wrongly.}
\vspace{-2mm}
\label{fig:qual}
\end{figure}

%% file: tables/qual_macros.tex
\providecommand{\qok}[1]{#1~\textcolor{green!50!black}{$\checkmark$}}
\providecommand{\qno}[1]{#1~\textcolor{red!70!black}{$\times$}}
\providecommand{\qualrows}[6]{%
\begin{tabular}[t]{@{}l@{\hspace{5pt}}l@{}}
TinyLLaVA-1.4B & #1 \\
Baseline+$(S_1,S_2)$ & #2 \\
Baseline+$(S_0,S_1,S_2)$ & #3 \\
Baseline(pruned)+$(S_0,S_1,S_2)$ & #4 \\
Recurrent LM+ViT+$(S_1,S_2)$ & #5 \\
RR-VLM (ours) & #6 \\
\end{tabular}}
\providecommand{\qualrow}[5]{%
\raisebox{\dimexpr\ht\strutbox-\height\relax}{\includegraphics[width=\linewidth]{#1}} &
\textbf{#2}\newline #3\newline #4 &
#5 \\}

%% file: sections/07_discussion.tex
\vspace{-6pt}%
\section{Conclusion}\label{sec:discussion}\label{sec:conclusion}
\vspace{-5pt}%
We introduced the first prune-then-loop retrofit of a pretrained VLM evaluated on VQA. The retrofitted model outperforms the unpruned baseline while storing 23.3\% fewer language-model parameters, with the gains concentrated on reasoning-oriented benchmarks. This gain is due to the recurrence itself, as the loop accounts for a significant improvement over the pruned single-pass baseline. At the same time, the recovery requires the text-only training stage, which yields the best performance among all evaluated training strategies. Finally, retrofitting recurrence into a pretrained VLM outperforms adding a vision encoder to a pretrained recurrent language model, which makes the retrofit the best performing route to a recurrent VLM among all evaluated baselines.

%% file: sections/appendix.tex
\setlength{\textfloatsep}{9pt plus 2pt minus 3pt}
\setlength{\floatsep}{7pt plus 2pt minus 2pt}
\setlength{\intextsep}{9pt plus 2pt minus 3pt}
\renewcommand{\topfraction}{0.95}
\renewcommand{\bottomfraction}{0.95}
\renewcommand{\textfraction}{0.05}
\renewcommand{\floatpagefraction}{0.85}
\setlength{\abovecaptionskip}{5pt}
\setlength{\belowcaptionskip}{2pt}

\clearpage
\section{Appendix}
\subsection{Training hyperparameters}\label{app:protocol}
Every stage uses Muon \citep{muon2024} for the matrix parameters and AdamW for all other
parameters, with weight decay $10^{-4}$, gradient clipping at 1.0 and a
warmup-stable-decay schedule with 5\% warmup. In $S_{0}$ the learning rates are
$5{\cdot}10^{-4}$ for Muon and $2.5{\cdot}10^{-5}$ for AdamW, the batch holds 128
sequences of 2,048 tokens, and the learning rate decays to zero over the last 35\% of
the steps. The 60B run of our model takes 228,874 steps on 16 GPUs, and the 7.5B run of
the ablations takes 28,611 steps. In $S_{1}$ and $S_{2}$ the learning rates are
$10^{-3}$ for Muon and $5{\cdot}10^{-5}$ for AdamW, the learning rate decays to zero
over the last 60\% of the steps, and both stages run on 4 GPUs. $S_{1}$ trains the
projector for 6,426 steps with a batch of 96 image-text pairs, and $S_{2}$ trains the
whole model for 6,408 steps with a batch of 128, one pass over the data in both stages
with the interleaved text batches included in the step counts.
The distillation weights and temperatures of each stage are given in
Sec.~\ref{sec:setup:recovery}. The Parcae adapter $A_{\mathrm{pc}}$ of Table~\ref{tab:ablation} uses an
additional learning rate of $10^{-3}$ for its adapter parameters, and the per-layer
looping variants of Table~\ref{tab:ablation}(c) share the recipe of the 7.5B cells. %
The converted recurrent VLM keeps the 4/8/4 geometry of our model and is trained with
its own trainer with the same learning rates and a batch of 128 in both stages.

\subsection{Distillation placement at the 7.5B budget}\label{app:kd75}
Table~\ref{tab:kd75} applies the distillation term of Eq.~\ref{eq:kd} in the stages $S_0$, $S_1$ and $S_2$, in $S_0$ only, or in $S_0$ and $S_2$ with weight $\lambda{=}0.50$ or $\lambda{=}0.25$, on the model with the $4/6/4$ layer allocation and $r{=}6$ at the 7.5B $S_0$ budget.

\input{tables/tab_kd75}
\FloatBarrier

\subsection{KV-cache sharing during training}\label{app:kvtrain}
Table~\ref{tab:kvtrain} trains the model with the $4/8/4$ layer allocation of our ablations at the 7.5B $S_0$ budget with the keys, or the keys and values, shared across passes (Sec.~\ref{sec:method:retrofit}).

\input{tables/tab_kvtrain}
\FloatBarrier

\subsection{Text interleaving in \texorpdfstring{$S_2$}{S2}}\label{app:interleave}
Stage $S_2$ interleaves one text-only batch after every $K$ multimodal batches. Table~\ref{tab:interleave} varies $K$ on the 60B model without text distillation, %
which is the VL-KD only row of Table~\ref{tab:kd}. Removing the text batches costs $0.69$ pp, and $K{=}2$ or $K{=}6$ cost $0.20$ pp and $0.07$ pp relative to $K{=}4$.

\input{tables/tab_interleave}
\FloatBarrier

\subsection{Early exit}\label{app:ttc}
To reduce inference cost, the step-wise generation of output logits (Eq.~\ref{eq:dynamics}) enables an adaptive halting criterion \citep{schuster2022calm,geiping2025huginn} that terminates the forward pass dynamically at step
\begin{equation}
k^{\star} \;=\; \min\{\,k\ge 2 : \operatorname{KL}\!\big(\operatorname{softmax}(y_{k})\,\big\|\,\operatorname{softmax}(y_{k-1})\big)<\epsilon\,\},
\label{eq:exit}
\end{equation}
where the threshold $\epsilon$ defines the accuracy-compute tradeoff.
In the read-out-only form every pass is still computed and the token is read from pass
$k^{\star}$. In the compute-skipped form the passes after $k^{\star}$ are not computed.
Table~\ref{tab:ttcexit} reports both forms on GQA, each against the same model without
exit ($\Delta$).

\input{tables/tab_ttcexit}
\FloatBarrier

\subsection{Qualitative examples}\label{app:qual}
Figures~\ref{fig:qual_app} and~\ref{fig:qual_app2} show two further items from each of the four reasoning-oriented benchmarks on which our model answers correctly and the three trained non-recurrent baselines of Table~\ref{tab:main}, Baseline+$(S_1,S_2)$, Baseline+$(S_0,S_1,S_2)$ and Baseline(pruned)+$(S_0,S_1,S_2)$, answer wrongly, together with the answer of every model in Table~\ref{tab:main}. A check mark denotes the correct option.

\input{tables/fig_qual_app}
\input{tables/fig_qual_app2}
\FloatBarrier

%% file: tables/tab_kd75.tex
\begin{table}[htbp]
\caption{\textbf{Placement of the distillation term at the 7.5B $S_0$ budget.} Each row shows in which training stages we apply the distillation term and with which weight $\lambda$, on the $4/6/4$ layer allocation with $r{=}6$. Seed 42. These are separate runs from the $4/6/4$ row of Table~\ref{tab:ablation}(b).}
\label{tab:kd75}
\centering
\footnotesize
\begin{tabular*}{0.6\textwidth}{@{\extracolsep{\fill}}lr@{}}
\toprule
\textbf{Distillation} & \textbf{Avg} \\
\midrule
$S_0,S_1,S_2$ (every stage) & 53.53 \\
$S_0$ only & 54.57 \\
$S_0,S_2$, $\lambda{=}0.50$ & 54.45 \\
$S_0,S_2$, $\lambda{=}0.25$ & \textbf{55.62} \\
\bottomrule
\end{tabular*}
\end{table}

%% file: tables/tab_kvtrain.tex
\begin{table}[htbp]
\caption{\textbf{KV-cache sharing across passes during training.} The model is trained with the
sharing active, at the 7.5B $S_0$ budget and the $4/8/4$ layer allocation of our ablations.
The Cache column gives the KV cache per token as a multiple of the base model's cache.
K-sharing shares the keys across passes, KV-sharing the keys and the values. Seed 42.}
\label{tab:kvtrain}
\centering
\scriptsize
\setlength{\tabcolsep}{0.5pt}
\begin{tabular*}{\textwidth}{@{\extracolsep{\fill}}@{}lrw{c}{3.1em}w{c}{3.1em}w{c}{3.1em}w{c}{3.1em}w{c}{3.1em}w{c}{3.1em}w{c}{3.1em}w{c}{3.1em}w{c}{3.1em}r@{}}
\toprule
\textbf{\boldmath Setting} & \textbf{\boldmath Cache} & \textbf{\boldmath GQA} & \textbf{\boldmath POPE} & \textbf{\boldmath TVQA} & \textbf{\boldmath SQA} & \textbf{\boldmath MMB} & \textbf{\boldmath RWQA} & \textbf{\boldmath SEED} & \textbf{\boldmath AI2D} & \textbf{\boldmath CV2D} & \textbf{\boldmath Avg} \\
\midrule
KV-sharing & $0.727\times$ & 57.6 & 86.1 & 44.5 & 51.1 & 50.0 & 43.8 & 57.8 & 37.3 & 47.1 & 52.81 \\
K-sharing & $1.64\times$ & 58.5 & 86.6 & 44.7 & 53.5 & 56.4 & 48.5 & 61.8 & 39.1 & 46.1 & 55.02 \\
no sharing & $2.55\times$ & 59.7 & 86.5 & 45.1 & 57.3 & 55.1 & 46.7 & 62.8 & 41.1 & 47.5 & 55.75 \\
\bottomrule
\end{tabular*}
\end{table}

%% file: tables/tab_interleave.tex
\begin{table}[htbp]
\caption{\textbf{Text interleaving in $S_2$.} Number of multimodal batches $K$ per text batch during $S_2$, on the 60B $S_0$ budget and the $4/8/4$ layer allocation without text distillation. The best value is in bold. Seed 42.}
\label{tab:interleave}
\centering
\footnotesize
\begin{tabular*}{0.6\textwidth}{@{\extracolsep{\fill}}lr@{}}
\toprule
\textbf{Interleaving} & \textbf{Avg} \\
\midrule
no text batches & 56.19 \\
$K{=}2$ & 56.68 \\
$K{=}6$ & 56.81 \\
$K{=}4$ (our setting) & \textbf{56.88} \\
\bottomrule
\end{tabular*}
\end{table}

%% file: tables/tab_ttcexit.tex
\begin{table}[htbp]
\caption{\textbf{Early exit (Eq.~\ref{eq:exit}) on GQA.} The retrofits use the 7.5B-token $S_{0}$ and are
trained at a fixed depth of 6 or with depth sampled at mean 6 or 16. The converted model at
mean depth 32 is built like the converted recurrent VLM of Table~\ref{tab:main} from the \citet{mcleish2025retrofitted} checkpoint
pretrained at that depth.
Pass read is the mean pass a generated token is read from. Passes run is the mean number of
block passes computed per generated token, with the first token counted at the cap
($^{\dagger}$: tokens after the first only). Settled@6 is the share of generated tokens whose
successive readouts have a KL divergence below $5{\cdot}10^{-3}$ at pass 6. Rows without
exit have no $\epsilon$ and no $\Delta$. The readouts of the fixed-depth model do not settle, so
the rule almost never fires on it (pass read 6.00 and 5.92). The values without exit differ from Tables~\ref{tab:main} and~\ref{tab:depth} because this evaluation ran without image padding.}
\label{tab:ttcexit}
\begin{center}
\footnotesize
\setlength{\tabcolsep}{3pt}
\begin{tabular*}{\textwidth}{@{\extracolsep{\fill}}llrrrrrrr@{}}
\toprule
\textbf{\boldmath Model} & \textbf{\boldmath Variant} & \textbf{\boldmath $\epsilon$} & \textbf{\boldmath Cap} & \textbf{\boldmath GQA} & \textbf{\boldmath $\Delta$} & \textbf{\boldmath \shortstack[r]{Pass\\read}} & \textbf{\boldmath \shortstack[r]{Passes\\run}} & \textbf{\boldmath \shortstack[r]{Settled\\@6}} \\
\midrule
sampled, mean 6 & no exit & -- & 6 & 58.92 & -- & 6.00 & 6.00 & 98.4\% \\
 & read-out only & $5{\cdot}10^{-3}$ & 6 & 58.53 & $-0.39$ & 2.81 & 6.00 &  \\
 & compute-skipped & $5{\cdot}10^{-3}$ & 6 & 58.31 & $-0.61$ & 2.79 & 3.75 &  \\
 & compute-skipped & $10^{-3}$ & 6 & 58.76 & $-0.16$ & 3.40 & 3.89 &  \\
\midrule
fixed, 6 & no exit & -- & 6 & 59.41 & -- & 6.00 & 6.00 & 3.7\% \\
 & read-out only & $5{\cdot}10^{-3}$ & 6 & 59.41 & $+0.00$ & 6.00 & 6.00 &  \\
 & compute-skipped & $5{\cdot}10^{-3}$ & 6 & 59.41 & $+0.00$ & 5.92 & 5.92 &  \\
\midrule
sampled, mean 16 & no exit & -- & 16 & 59.06 & -- & 16.00 & 16.00 & -- \\
 & read-out only & $5{\cdot}10^{-3}$ & 16 & 58.40 & $-0.67$ & 3.15 & 16.00 &  \\
 & compute-skipped & $5{\cdot}10^{-3}$ & 16 & 58.55 & $-0.52$ & 2.28$^{\dagger}$ & 2.28$^{\dagger}$ &  \\
 & no exit & -- & 32 & 59.12 & -- & 32.00 & 32.00 &  \\
 & compute-skipped & $5{\cdot}10^{-3}$ & 32 & 58.71 & $-0.41$ & 2.28$^{\dagger}$ & 2.28$^{\dagger}$ &  \\
\midrule
converted, 8 & no exit & -- & 8 & 58.36 & -- & 8.00 & 8.00 & 71.4\% \\
 & read-out only & $5{\cdot}10^{-3}$ & 8 & 57.40 & $-0.96$ & 3.77 & 8.00 &  \\
 & compute-skipped & $5{\cdot}10^{-3}$ & 8 & 57.39 & $-0.98$ & 3.72 & 4.72 &  \\
\midrule
converted, mean 32 & no exit & -- & 32 & 59.04 & -- & 32.00 & 32.00 & 77.3\% \\
 & read-out only & $5{\cdot}10^{-3}$ & 32 & 58.27 & $-0.77$ & 3.97 & 32.00 &  \\
 & compute-skipped & $5{\cdot}10^{-3}$ & 32 & 58.20 & $-0.83$ & 3.93 & 14.70 &  \\
\bottomrule
\end{tabular*}
\end{center}
\end{table}

%% file: tables/fig_qual_app.tex
\input{tables/qual_macros}
\begin{figure}[htbp]
\centering
\scriptsize
\begin{tabular}{@{}p{0.16\textwidth}>{\raggedright\arraybackslash}p{0.33\textwidth}p{0.33\textwidth}@{}}
\toprule
\qualrow{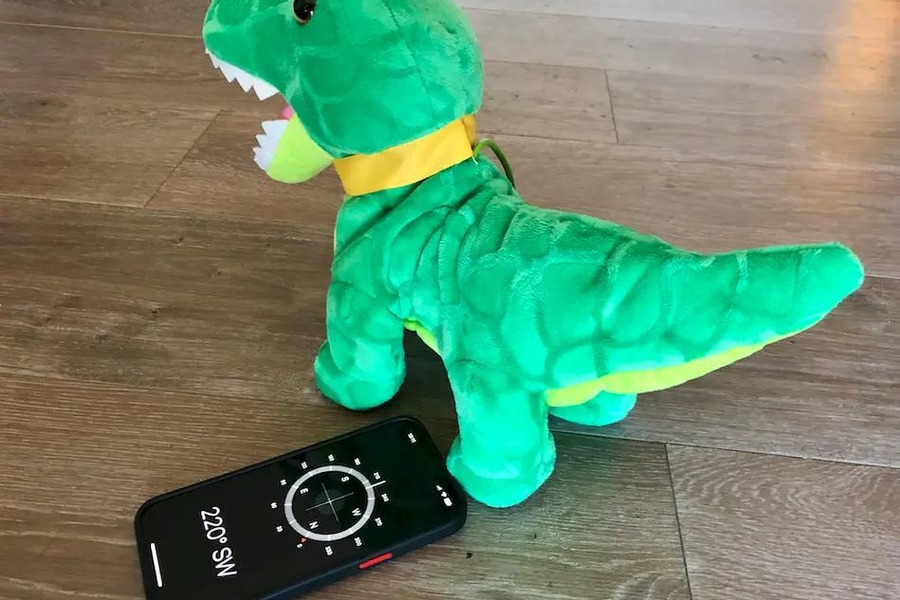}{RealWorldQA}
{Given the picture, in which cardinal direction is the dinosaur facing?}
{A.~North. \quad B.~South. \quad C.~East. \quad D.~West}
{\qualrows{\qno{D}}{\qno{A}}{\qno{A}}{\qno{D}}{\qok{C}}{\qok{C}}}
\midrule
\qualrow{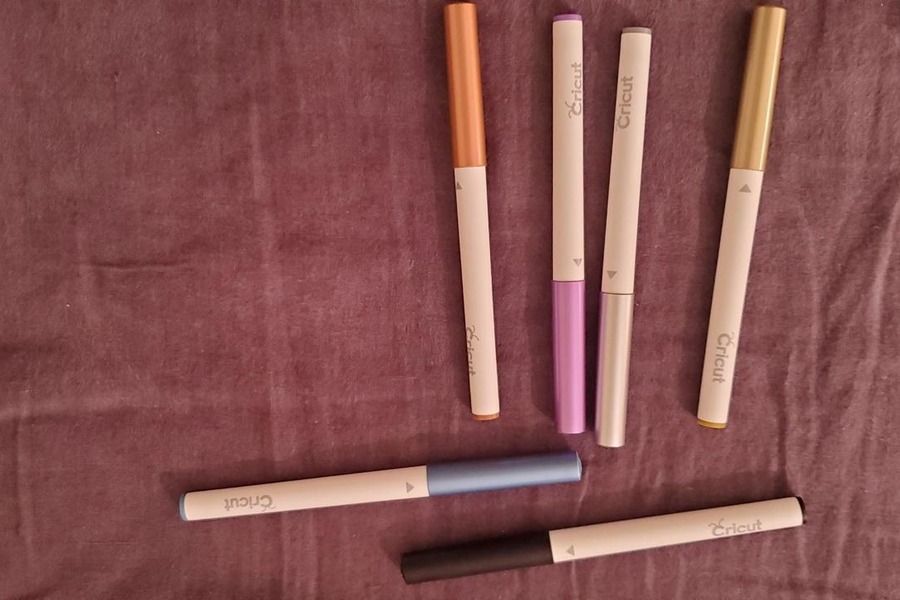}{RealWorldQA}
{Which direction is the blue cap pens arrow pointing?}
{A.~Pointing up. \newline B.~Pointing to the right. \newline C.~Pointing to the left.}
{\qualrows{\qno{C}}{\qno{C}}{\qno{C}}{\qno{C}}{\qno{C}}{\qok{B}}}
\midrule
\qualrow{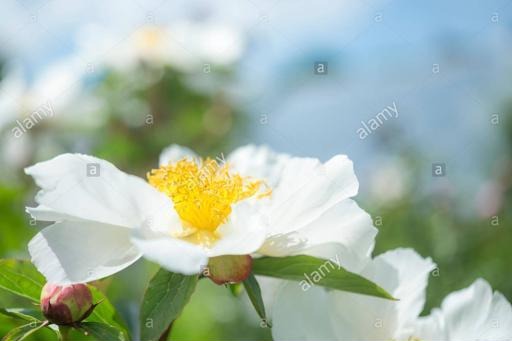}{SEED-Bench}
{What is the relative position between the leaf and the flower?}
{A.~The leaf is above the flower \newline B.~The leaf is on the right of the flower \newline C.~The leaf is on the left of the flower \newline D.~The leaf is below the flower}
{\qualrows{\qno{A}}{\qno{A}}{\qno{A}}{\qno{A}}{\qno{A}}{\qok{D}}}
\midrule
\qualrow{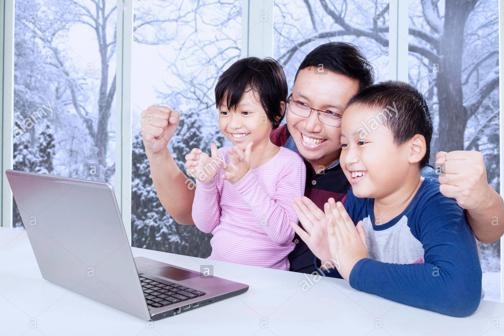}{SEED-Bench}
{Is the laptop above, below, or next to the family in the image?}
{A.~Next~to \quad B.~Below \quad C.~Above \quad D.~In~front~of}
{\qualrows{\qok{D}}{\qno{C}}{\qno{C}}{\qno{C}}{\qok{D}}{\qok{D}}}
\bottomrule
\end{tabular}
\caption{\textbf{Qualitative examples (RealWorldQA, SEED-Bench).} Two items per benchmark with the answer of every model in Table~\ref{tab:main}. A check mark denotes the correct option.}
\label{fig:qual_app}
\end{figure}

%% file: tables/fig_qual_app2.tex
\begin{figure}[H]
\centering
\scriptsize
\begin{tabular}{@{}p{0.16\textwidth}>{\raggedright\arraybackslash}p{0.33\textwidth}p{0.33\textwidth}@{}}
\toprule
\qualrow{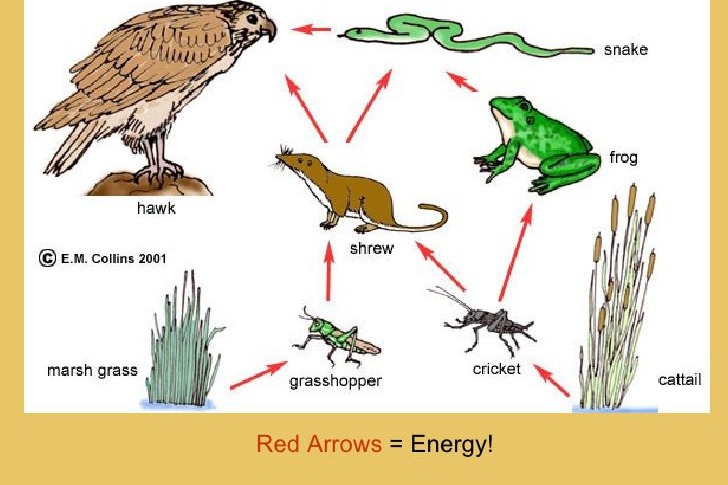}{AI2D}
{Which is a secondary consumer?}
{A.~shrew \quad B.~grasshopper \quad C.~cricket \quad D.~grass}
{\qualrows{\qno{B}}{\qno{B}}{\qno{B}}{\qno{B}}{\qno{B}}{\qok{A}}}
\midrule
\qualrow{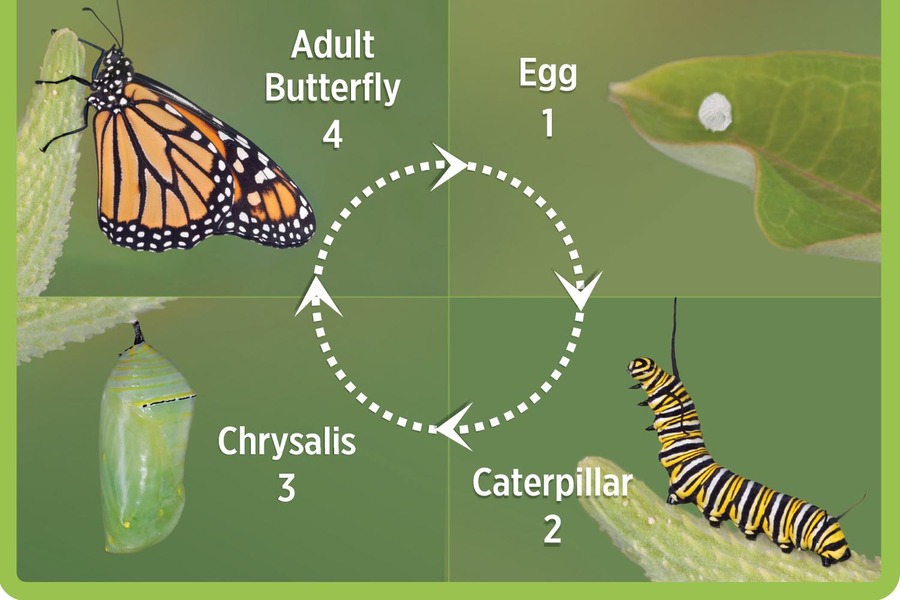}{AI2D}
{What does stage the number 3 depict?}
{A.~Adult \quad B.~Egg \quad C.~Chrysalis \quad D.~Caterpillar}
{\qualrows{\qno{B}}{\qno{B}}{\qno{B}}{\qno{D}}{\qno{B}}{\qok{C}}}
\midrule
\qualrow{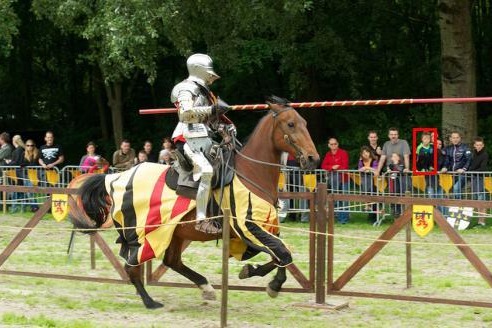}{CV-Bench-2D}
{Considering the relative positions of the person (annotated by the red box) and the horse in the image provided, where is the person (annotated by the red box) located with respect to the horse?}
{A.~left \quad B.~right}
{\qualrows{\qno{A}}{\qno{A}}{\qno{A}}{\qno{A}}{\qno{A}}{\qok{B}}}
\midrule
\qualrow{}{CV-Bench-2D}
{Considering the relative positions of the cutlery and the range in the image provided, where is the cutlery located with respect to the range?}
{A.~left \quad B.~right}
{\qualrows{\qno{A}}{\qno{A}}{\qno{A}}{\qno{A}}{\qno{A}}{\qok{B}}}
\bottomrule
\end{tabular}
\caption{\textbf{Qualitative examples (AI2D, CV-Bench-2D).} Two items per benchmark with the answer of every model in Table~\ref{tab:main}. A check mark denotes the correct option.}
\label{fig:qual_app2}
\end{figure}